# PLCNet: Patch-wise Lane Correction Network for Automatic Lane Correction in High-definition Maps

Haiyang Peng, Yi Zhan, Benkang Wang, Hongtao Zhang

*Abstract*—In High-definition (HD) maps, lane elements constitute the majority of components and demand stringent localization requirements to ensure safe vehicle navigation. Vision lane detection with LiDAR position assignment is a prevalent method to acquire initial lanes for HD maps. However, due to incorrect vision detection and coarse camera-LiDAR calibration, initial lanes may deviate from their true positions within an uncertain range. To mitigate the need for manual lane correction, we propose a patch-wise lane correction network (PLCNet) to automatically correct the positions of initial lane points in local LiDAR images that are transformed from point clouds. PLCNet first extracts multi-scale image features and crops patch (ROI) features centered at each initial lane point. By applying ROIAlign, the fix-sized ROI features are flattened into 1D features. Then, a 1D lane attention module is devised to compute instance-level lane features with adaptive weights. Finally, lane correction offsets are inferred by a multi-layer perceptron and used to correct the initial lane positions. Considering practical applications, our automatic method supports merging local corrected lanes into global corrected lanes. Through extensive experiments on a self-built dataset, we demonstrate that PLCNet achieves fast and effective initial lane correction.

## I. INTRODUCTION

High-definition (HD) maps serve as an essential component in furnishing geometric and semantic map information to facilitate autonomous driving systems. The HD map information contains positions and attributes of massive static traffic objects, e.g., lane, road marking and traffic light. In general, HD maps are created from labor-intensive annotations on LiDAR point clouds combined with front-view images, high-resolution aerial images or reconstructed point cloud images by professional surveying and mapping engineers. Due to the time-consuming and costly nature of manual annotation, it is imperative to automate the HD map annotation process. A common practice is to replace hand-crafted feature extraction with automatic methods. With the rapid development of deep learning, many automatic methods have been developed to extract specified HD map elements, such as lane, boundary, road marking and traffic light [1]. Among these elements, lane and boundary detection have garnered significant attention.

In recent literature, earlier approaches primarily focused on detecting lanes or boundaries in front-view images [2]-[4],

Haiyang Peng, Yi Zhan, Benkang Wang and Hongtao Zhang are with Uimobi Technology, Zhejiang, China. Yi Zhan is the corresponding author: yi.zhan@uisee.com

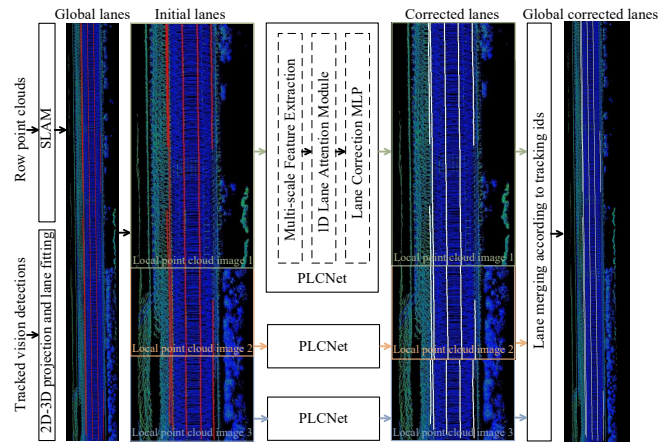

Fig. 1: Automatic initial lane correction via PLCNet. Global and initial lanes are marked by red color. Corrected and global corrected lanes are marked by white color. Please refer to section III for more details.

aerial images [5]-[6], and point cloud images [7]. Subsequently, post-processing techniques such as LiDAR position assignment, tracking, fitting and stitching are used to form the holistic HD maps. In contrast, more recent methods [8]-[9] aim to construct end-to-end HD maps directly on vehicle hardware by fusing local bird's-eye-view (BEV) information from LiDAR and camera, thereby considering both geometric and semantic features. In practical application, the following three cases may lead to produce deviated lanes on front-view images: (1) due to image resolution limitations, accurately detecting lane points at a considerable distance from the ego vehicle poses a significant challenge; (2) factors such as vehicle vibrations, sensor degradation and environmental changes may render the raw camera-LiDAR calibration inaccurate, leading to misaligned lanes after 2D-3D projection; (3) when lanes are occluded by other objects, their 3D projection coordinates lie on the objects rather than the lanes themselves. Such deviated lanes are not acceptable in high-level autonomous driving systems where safety is of paramount importance. The most straightforward solution is to conduct manual correction on the initial lanes with possible deviation. However, this approach is not only time-consuming but also susceptible to human error.

To address the above issues, this paper proposes a patch-wise lane correction network (PLCNet) to automatically correct the initial lanes for HD maps. PLCNet leverages the strengths of MLP [10], CNN [11] and attention [12] and has a simple yet effective network architecture. As depicted in Fig. 1, PLCNet first uses a multi-scale CNN backbone to get rich feature maps on point cloud images, which are cropped from 3D reconstruction maps. We gather patch-wise features centered at each initial lane point. Then a 1D lane attention module is used to extract instance-level lane features. Next, a simple MLP infers the corresponding lane correction offsets, which is in accordance with the process of manual correction. The corrected lanes are obtained by adding the initial lanes with offsets. Finally, global corrected lanes in HD maps are constructed by merging the corrected lanes with same global instance numbers sourced from vision lane tracking identities (IDs). To verify the effectiveness of PLCNet, we apply it to complete automatic lane correction under self-built urban, highway and factory scenes. Furthermore, we compare PLCNet with a modified Deepsnake method [13] to showcase its superior performance. The contributions of this work are listed as follows:

- To the best of our knowledge, PLCNet is the first end-to-end model to solve the issue of automatic initial lane correction for HD maps, with potential applications in related domains in the future.
- PLCNet capitalizes on the benefits of MLP, CNN and attention and integrates them in a unified one-stage network. In addition, PLCNet considers multi-scale contextual information of each lane and inherently supports instance-level lane correction without any iterative evolution.
- We created a self-built lane correction dataset encompassing urban, highway and factory scenes. Through both qualitative and quantitative evaluations on this dataset, we demonstrate the superiority of PLCNet compared to Deepsnake.

## II. RELATED WORKS

### A. Vision Lane and Boundary Detection Methods

In general, the construction of HD maps involves an initial detection of line-shaped elements (e.g., lane and boundary) from specified image sources followed by the application of post-processing algorithms to generate global maps. For the front-view images, Neven *et al*. [2] propose a LaneNet architecture that employs a segmentation-then-clustering approach to produce lane-wise points. Concurrently, H-Net learns a robust perspective transformation to fit these lane points. LaneATT [3] incorporates an attention mechanism to extract global features from local ones and conducts anchor-based lane detection. BézierLaneNet [4] further investigates the use of parametric Bézier curves for improved lane modeling. For aerial images, Xu *et al*. [5] introduce a boundary dataset in this domain and evaluate several baseline models on this dataset. In [6], a boundary graph is computed by predicting potential boundary vertices and their corresponding adjacency matrix. Liang *et al*. [7] initially employ a fully convolutional network to obtain detection, endpoint, and direction maps from point cloud images and overhead images. Subsequently, these maps are fed into a convolutional recurrent network to directly infer structured boundary polylines.

### B. End-to-end Local HD Map Construction Methods

In traditional vision-based lane and boundary detection methods, the derived results lack absolute positional information. To address this issue, LiDAR data is employed to assign position coordinates. However, recent approaches have shifted towards constructing local HD maps in an end-to-end fashion. To fully utilize multi-modal data, they exploit the feature-level BEV fusion of LiDAR and camera instead of point-level LiDAR position assignment. Similar to [2], HDMapNet [8] adopts the segmentation-then-clustering scheme to obtain vectorized lane and boundary elements. VectorMapNet [9] removes the heavy post-processing clustering and manipulates an auto-regressive fashion to predict sequential vertices of each line-shaped element. Despite promoting the inference speed, the feature misalignment of LiDAR and camera has not been tackled, which often occurs in multi-modal data fusion.

### C. Contour-based Instance Segmentation Methods

Since there is no existing lane correction method, we opt to compare our proposed PLCNet with the most relevant approach, Deepsnake. Cootes *et al*. [14] conduct the seminal work to apply snake algorithms in image segmentation, which optimize an energy function with respect to initial contour vertices. In fact, Deepsnake is a contour-based instance segmentation method that first using a detector to produce initial proposals and then introducing the circular convolution to predict the offsets between the initial boundary and ground truth boundary. It is worth noting that Deepsnake focuses on closed-shape object segmentation and requires several evolution iterations for distant boundary regression. For the sake of fair comparison, we modify Deepsnake and disable its augmentation strategy as illustrated in section IV-E.

## III. AUTOMATIC INITIAL LANE CORRECTION

To mitigate the burden of manual correction, this paper proposes an automatic initial lane correction pipeline for HD maps. As shown in Fig.1, this pipeline includes three primary steps: (1) data acquisition of point cloud images and initial lanes; (2) taking the data as input and outputting corrected lanes via PLCNet; (3) merging the corrected lanes into global corrected lanes.

### A. Point Cloud Image and Initial Lane Acquisition

In our study, autonomous vehicles are furnished with an array of sensors, including LiDAR, camera, GPS, wheel and IMU, to gather multi-modal data across diverse driving environments. The presence of dynamic objects, such as moving vehicles and pedestrians, introduces noises into the LiDAR-based 3D reconstruction map. Hence, a 3D detector [15] is applied to identify and filter out these noisy points at the point cloud level. Then, we choose graph optimization-based Cartographer [16] and ground optimization-based LeGO-LOAM [17] as the initial SLAM algorithms. Next, the initial results from the two algorithms are fused with IMU to obtain robust 3D reconstruction map and trajectory. Point cloud images can be acquired through projecting the reconstruction map from $X$-$Y$ plane into $x$-$y$ image plane with a specified pixel resolution $R$. To save computational resources, we sample a fix-sized pixel region ($H \times W$) along the trajectory at a sampling interval $S$ and obtain $N$ local point cloud images $\{I_i\}_{i=1}^{N}$ ($I_i$ is the $i$th image). Simultaneously, the left-bottom absolute coordinates $\{(X_i^{lb}, Y_i^{lb})\}_{i=1}^{N}$ of sampled regions are recorded for image to geographical coordinate transformation. For the acquisition of initial lanes, we first use LaneNet [2] to detect lanes on the front-view images and obtain their tracking IDs by [18]. Then, camera-LiDAR calibration and vehicle trajectory are applied to complete 2D-3D projection. Next, the projected lanes with same tracking IDs are fitted into global lanes. In the end, the region sampling method is also used to crop local initial lanes $\{L_i\}_{i=1}^{N}$ with tracking IDs on the global lanes.

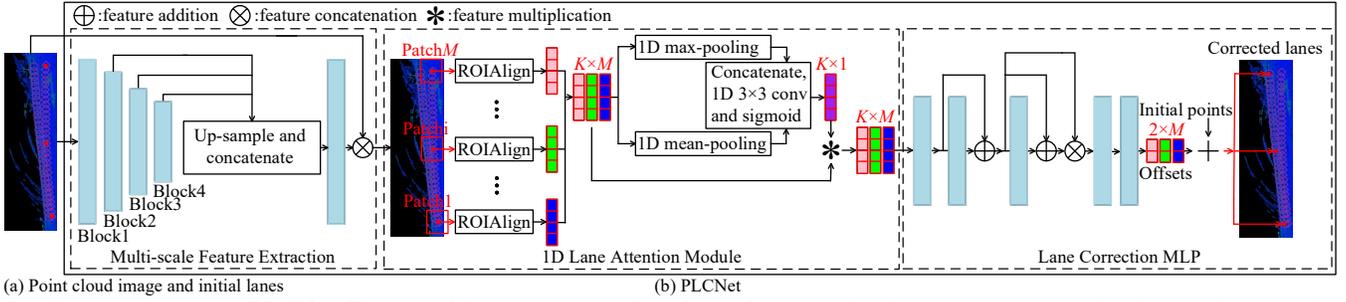

(a) Point cloud image and initial lanes   (b) PLCNet

Fig. 2: An overview of PLCNet. The initial lanes and ground truths in the point cloud image are marked by diamond and circle, respectively. (a) PLCNet takes a point cloud image and initial lanes as input; (b) PLCNet includes the multi-scale feature extraction, 1D lane attention module and instance-level lane correction MLP. To be concrete, we first use a light-weight EfficientNet-B0 backbone to extract multi-scale features. Then we crop image patch (ROI) features centered at each initial lane point and employs ROIAlign to output fix-sized features (size: $K \times M$). Thereafter, the 1D lane attention module aggregates lane features (size: $K \times M$) and a simple MLP with five 1D 1×1 convolutions (dimension: 64D) is utilized to infer instance-level lane correction offsets (size: $2 \times M$). Finally, corrected lanes are obtained by adding the offsets and initial lanes. For better view, only one lane instance is shown in the point cloud image.

### B. PLCNet

We design PLCNet through a concise and elegant paradigm as overviewed in Fig. 2, which contains three parts: (1) multi-scale feature extraction, (2) 1D lane attention module and (3) lane correction MLP.

**Multi-scale Feature Extraction.** After the collection of point cloud images, the traditional lane correction requires manual comparison between initial lanes and their adjacent image features. As mentioned before, replacing the hand-crafted feature extraction with deep learning is pivotal to automate the lane correction. In pursuit of high accuracy and performance, we utilize the light-weight EfficientNet-B0 [19] as the feature extraction backbone. The backbone is comprised of convolutional Blocks1-4 for simplicity, which correspond to $2^1 \times \sim 2^4 \times$ down-sampled feature maps. To aggregate multi-scale feature information, the feature maps of Blocks2-4 are up-sampled to the same input size $H \times W$ and then stacked together. Following this, a final 2D convolution is employed to derive the binary lane segmentation map. To better emulate the process of manual comparison, the segmentation map is concatenated with the input image $I_i$ into a 4D feature tensor $F_i$. The overall extraction procedure is formulated as:

$$F_i = C\left(\left\{I_i, Conv2D_{1\times1}\left(C\left(\left\{Up_j(Block_j(I_i))\right\}_{j=2}^{4}\right)\right)\right\}\right) \quad (1)$$

where $Up_j(\cdot)$ is the $j$th up-sampling operation, and $\{Up_j(Block_j(I_i))\}_{j=2}^{4}$ denotes the set of up-sampled Blocks2-4 features. $C(\cdot)$ is the feature concatenation.

**1D Lane Attention Module.** In addition to the multi-scale features, local lane features with rich contextual information are also essential for predicting accurate correction offsets. Specifically, we crop a $P \times P$-sized patch centered at each initial lane point on $F_i$. Formally, we use $L_i = \{l_{i,j}\}_{j=1}^{N_i}$ to represent initial lane instance set with the number of $N_i$ in point cloud image $I_i$, where $l_{i,j} = \{v_k \in \mathbb{R}^2\}_{k=1}^{M}$ is the $j$th lane that contains $M$ B-spline interpolation points $v_k = (x_k, y_k)$. Since these patches may exceed the image range ($H \times W$), we only crop the valid patch area. The resulting patch size may not be $P \times P$, which is adverse to integrate the patch-wise features into a tensor. Therefore, ROIAlign [20] with flattening is used to output $K \times 1 = (P \times P \times 4) \times 1$ feature map on arbitrary-sized valid patch. Consequently, we can obtain the integrated feature tensor $T_i$ of size $N_i \times K \times M$ through traversing all initial lane instances. After obtaining $T_i$, we design a 1D lane attention module to extract instance-level lane features. This module first applies max-pooling and mean-pooling operations in the $M$ dimension of $T_i$ for each lane instance. Thereafter, a 1D $3 \times 3$ convolution followed by a sigmoid function is used to compute instance-level feature weight. According to the principle of attention mechanism, $T_i$ will be multiplied by the weight to obtain the final features $A_i$. The attention process is formulated as:

$$A_i = T_i * \sigma\left(Conv1D_{3\times3}\left(C\left(\{\max(T_i), \text{mean}(T_i)\}\right)\right)\right) \quad (2)$$

where $\max(\cdot)$ and $\text{mean}(\cdot)$ denote the max-pooling and mean-pooling operation, respectively. $\sigma(\cdot)$ is the sigmoid function.

**Lane Correction MLP.** In the object detection field, MLP is often used to regress bounding box offsets. Similarly, we leverage MLP to infer instance-level lane correction offsets $O_i = \{o_{i,j}\}_{j=1}^{N_i} = \{\{o_{i,j,k} = (\Delta x_{i,j,k}, \Delta y_{i,j,k}) \in \mathbb{R}^2\}_{k=1}^{M}\}_{j=1}^{N_i}$ after obtaining the feature tensor $A_i$. For constructing a fully convolutional PLCNet, we replace each MLP layer by an equivalent 1D $1 \times 1$ convolution. To prevent the risk of over-fitting, the MLP has only five 1D $1 \times 1$ convolutions and the last convolution is not followed by any activation function. Specific architecture can be found in Fig. 2. In the end, corrected lanes $C_i = \{c_{i,j}\}_{j=1}^{N_i}$ are obtained by adding the offsets and initial lanes. Thanks to the aforementioned multi-scale and adaptive lane features, PLCNet does not use any iterative evolution to tackle the challenge of far lane correction.

### C. Merging of Corrected Lanes

After obtaining the corrected lanes, they are converted from image coordinates to absolute coordinates via the recorded left-bottom region coordinates $\{(X_i^{lb}, Y_i^{lb})\}_{i=1}^{N}$. Then, we group the converted lanes according to the tracking IDs and smooth the grouped lanes. To be specific, the converted lanes with same tracking IDs are grouped into global corrected lanes. We utilize linear interpolation to smooth each global corrected lane with 100 interpolated points. For the convenience of subsequent evaluation, the same interpolation is applied to smooth the global corrected lanes and corresponding ground truth lanes. Fig. 3 illustrates the details of the merging process.

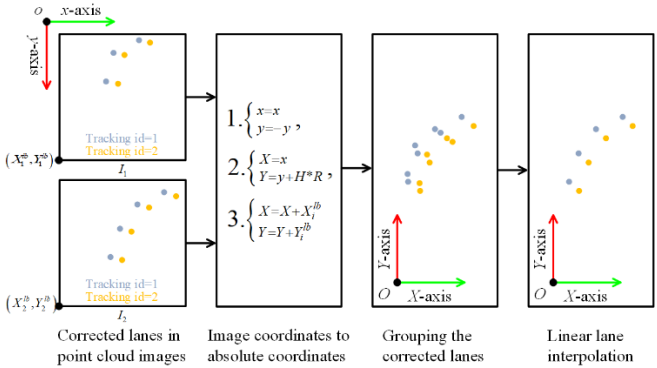

Fig. 3: Details of merging process. The image and geographical coordinate systems are represented by x-y and X-Y, respectively. $(X_i^{lb}, Y_i^{lb})$ is the left-bottom absolute coordinate of local point cloud image $I_i$. $H$ and $R$ are the image height and pixel resolution, respectively.

## IV. EXPERIMENTAL RESULTS AND DISCUSSIONS

### A. Dataset

Since there is no existing dataset for lane correction, this work built a self-built dataset with initial lanes and point cloud images under urban, highway and factory driving scenes. To be concrete, each point cloud image $I_i$ has the size of $H \times W = 2800 \times 1400$ and the pixel resolution of $R = 0.1\text{m} \times 0.1\text{m/pixel}$. As formulated in Eq. (3), the RGB values of the $t$th image pixel are rendered by its point cloud intensity $U_t$. The sampling interval $S$ on the trajectory is 40m. The method of obtaining initial lanes is mentioned in section III-A. To train PLCNet end-to-end, we collect accurate lane correction ground truths that are instance-level and have tracking IDs of matched initial lanes by human labelers. After extending each lane correction point to 1 pixel range, binary segmentation label can be collected as well. Fig. 4 shows one sample of point cloud image and binary segmentation label. In total, this dataset has 325 point cloud images, 4126 lane instances and covers 19.5km scene range. Data statistical result is listed in Tab. I.

$$\begin{aligned} r &= (U_t - U_{\min})/(U_{\max} - U_{\min}), \\ R &= 0, \\ G &= \lfloor r * 255 \rfloor, \\ B &= \lfloor (1-r) * 255 \rfloor \end{aligned} \quad (3)$$

where $U_{\min}$ and $U_{\max}$ are the minimum and maximum values in all point cloud intensities, respectively. $r$ is the ratio of G and B components. $\lfloor \cdot \rfloor$ is the rounding down operation.

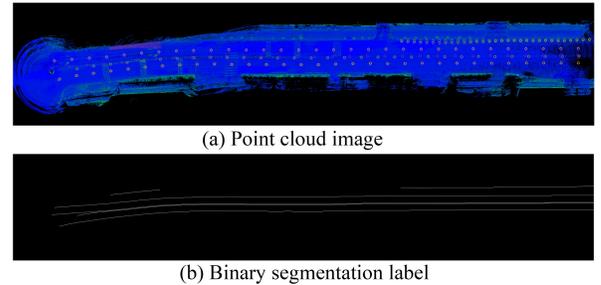

(a) Point cloud image

(b) Binary segmentation label

Fig. 4: One sample of point cloud image and binary segmentation label. The initial lanes and ground truths in the point cloud image are marked by diamond and circle, respectively. Same lane instance is identified by same color. Lane points are marked as 1, while the background points are marked as 0 in the binary segmentation label.

TABLE I: The statistical result of our self-built dataset.

|  | # Point cloud images | # Lane instances | Coverage area (km) |
|---|---|---|---|
| Urban scene | 223 | 3159 | 14 |
| Highway scene | 88 | 782 | 5 |
| Factory scene | 14 | 185 | 0.5 |

### B. Implementation Details

In our experiments, we train PLCNet end-to-end for 60 epochs with Adam [21]. The initial learning rate is 0.001 and it will be reduced to 0.0001 after 50 epochs. To speed up the network convergence, EfficientNet-B0 is pre-trained on ImageNet [22]. The training batch size is set as 2 and each sample is resized to $H \times W = 640 \times 320$. The whole dataset is split into training and testing datasets with a ratio of 3:2. The interpolated number $M$ is 32 and the patch size $P \times P$ is $6 \times 6$. PLCNet outputs a binary segmentation map from the multi-scale lane feature extraction and instance-level lane correction offsets from the MLP. We use focal loss and smooth-L1 loss applied in [13] to optimize the binary segmentation map and lane correction process, respectively. During inference, lane correction offsets are added by initial lanes to obtain corrected lanes. The corrected lanes are resized to the original scale and then merged into global corrected lanes with tracking IDs. We conduct all the experiments on a PC with i7-8700K CPU and RTX2080Ti GPU.

*C. Evaluation Metrics*

In our experiments, we use four evaluation metrics including two point-level metrics (i.e., smooth-L1 and L2 distances) for local coarse evaluation and two lane-level metrics (i.e., lane intersection over union (lane-IoU) and Chamfer distance (CD)) for holistic fine evaluation. All the metrics are averaged by the number of lane instances.

1) Point-level metrics: smooth-L1 and L2 distances are widely-used measurement between corrected and ground truth lanes. However, they independently compute distances point by point rather than consider the overall lane distance. The two formulae are given below:

$$d_{j,k} = \hat{v}_k - v_k \text{ where } \hat{v}_k \in \hat{l}_j, v_k \in l_j,$$
$$\text{smooth-L1}(d_{j,k}) = \begin{cases} 0.5 * d_{j,k}^2 & \text{if L1 norm}(d_{j,k}) < 1 \\ \text{L1 norm}(d_{j,k}) - 0.5 & \text{otherwise} \end{cases}, (4)$$
$$\text{L2}(d_{j,k}) = \text{L2 norm}(d_{j,k})$$

where $\hat{v}_k$ and $v_k$ are the $k$th points on the $j$th corrected lane $\hat{l}_j$ and ground truth lane $l_j$, respectively. $\text{L1 norm}(\cdot)$ and $\text{L2 norm}(\cdot)$ are L1 and L2 norms in mathematics, respectively.

2) Lane-level metrics: IoU is a standard semantic segmentation metric that computing the overlapping ratio of predicted segmentation map and ground truth label. To measure lane-level lane distance, we provide a new lane-IoU metric that first extends both corrected and ground truth points to $K$ pixels ($K \in \{1,2,3\}$) with label 1. Then, other background pixels are marked as label 0. In this way, we can construct binary mask maps and compute IoU lane by lane:

$$\text{lane-IoU}(\hat{m}_j, m_j) = \frac{|\hat{m}_j \cap m_j|}{|\hat{m}_j \cup m_j|} \quad (5)$$

where $\hat{m}_j$ is the extended mask of the $j$th corrected lane $\hat{l}_j$ and $m_j$ is the extended mask of the $j$th ground truth lane $l_j$. $\cap$ and $\cup$ are intersection and union operations, respectively. $|\cdot|$ computes the collected pixel number.

Since IoU is based on pixel computation, it may introduce quantization error. Therefore, we use CD metric to evaluate spatial distances between corrected and ground truth lanes. Note that we consider the mean bidirectional CD:

$$\text{CD}_{dir}(\hat{l}_j, l_j) = \frac{1}{|\hat{l}_j|} \sum_{\hat{v} \in \hat{l}_j} \min_{v \in l_j} \left( \text{L2 norm}(\hat{v} - v) \right),$$
$$\text{CD}(\hat{l}_j, l_j) = \frac{1}{2} \left( \text{CD}_{dir}(\hat{l}_j, l_j) + \text{CD}_{dir}(l_j, \hat{l}_j) \right) \quad (6)$$

where $\text{CD}_{dir}(\cdot)$ computes the bidirectional CD.

*D. Baseline Model*

We select Deepsnake [13] as the baseline model, which focuses on real-time instance segmentation for natural objects. It first detects initial object contour and then uses the proposed circular convolution blocks to infer the offsets between the initial and ground truth contours. Although the above workflow is similar to our lane correction, there are still some significant differences: (1) Deepsnake is a two-stage model for closed-shape object segmentation whose performance relying on detections. In contrast, PLCNet is a one-stage model and designed for open-shape lane correction; (2) PLCNet solves distant lane correction using multi-scale, patch-wise, and instance-level features instead of three time-consuming iterations in Deepsnake. To make fair network comparison, all the experiments are conducted under same input size, batch size and training optimizer without any data augmentation. We make two modifications for the lane correction task: (1) setting initial lanes as the initial contours and (2) only optimizing the center and smooth-L1 loss [13].

*E. Comparable Result on Local and Global Correction*

In this section, we evaluate PLCNet with the baseline model Deepsnake on the self-built lane correction dataset. Local and global correction under image and absolute coordinate systems are evaluated on pixel (p) and meter (m) levels, respectively. Fig. 5 and Tab. II provide the qualitative and quantitative comparison results of local corrected lanes, respectively. Tab. III lists the quantitative comparison results of global corrected lanes.

From Tab. II, it can be found that PLCNet outperforms Deepsnake on both point-level and lane-level metrics. Specifically, the performance margin of smooth-L1 and L2 is significant, which proves that PLCNet can predict accurate lane point correction. Thanks to the proposed 1D lane attention module, PLCNet focuses on instance-level lane correction and obtains better lane-IoU and CD. This result also indicates that the proposed circular convolution of Deepsnake is not applicable to open-shape lane correction. As shown in Fig. 5, PLCNet has better qualitative results than Deepsnake on urban, highway and factory scenes. In despite of using evolution strategy, Deepsnake does not infer adequate offsets for the initial lanes that are far from exact positions. In addition, we select some intricate scenes with unclear lane features and close lane instances to further demonstrate the effectiveness of PLCNet. From Fig. 5(d), it is observed that PLCNet can deal with these challenging cases and infers precise correction offsets. On the other hand, both methods are able to correct global initial lanes and the global corrected lanes of PLCNet are closer to the ground truths than that of Deepsnake, as listed in Tab. III. To sum up, PLCNet exhibits superior performance in terms of both local and global lane correction.

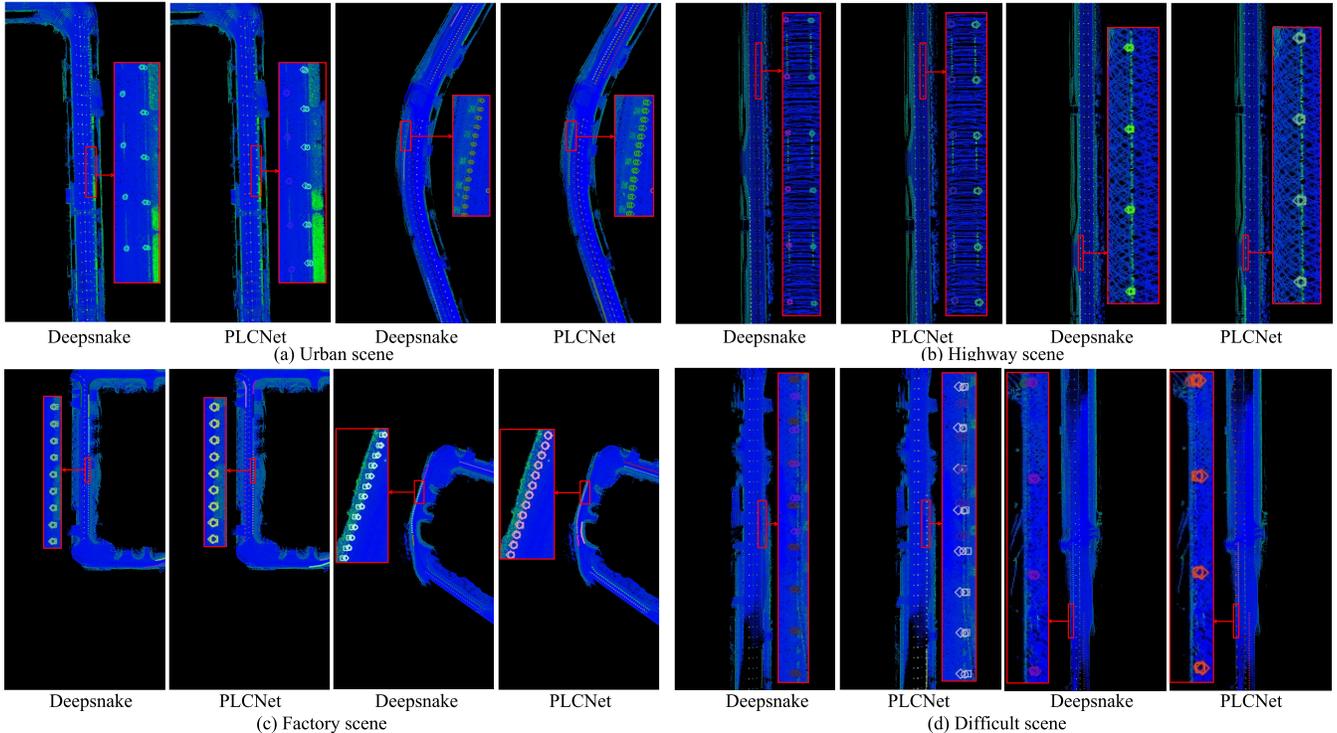

Fig. 5: The qualitative comparison results of local corrected lanes on urban, highway and factory scenes. The initial lanes and corrected lanes and ground truths are marked by diamond, square and circle, respectively. The intricate scene includes point cloud images with unclear lane features and close lane instances. More details within red rectangles are zoomed in.

TABLE II: The quantitative comparison results of local corrected lanes. 1p, 2ps and 3ps represent 1 pixel, 2 pixels and 3 pixels extension on each lane point, respectively.

| Methods | lane-IoU (%) ↑ | | | smooth-L1 ↓ (p) | L2 ↓ (p) | CD ↓ (p) |
|---|---|---|---|---|---|---|
| | 1 p | 2 ps | 3 ps | | | |
| Deepsnake | 21.9 | 51.1 | 65.0 | 0.148 | 0.574 | 1.339 |
| PLCNet | **23.5** | **54.8** | **68.6** | **0.040** | **0.257** | **1.135** |

TABLE III: The quantitative comparison results of global corrected lanes. The initial state means that all the initial lanes have not been corrected by any network.

| Methods | smooth-L1 ↓ (m) | L2 ↓ (m) | CD ↓ (m) |
|---|---|---|---|
| Initial state | 0.121 | 0.476 | 0.282 |
| Deepsnake | 0.111 | 0.425 | 0.234 |
| PLCNet | **0.093** | **0.360** | **0.207** |

### F. Ablation Studies

Since PLCNet consists of two proposed components, namely, 1D lane attention module and patch feature extraction, it is a need to validate their importance of constructing PLCNet. Therefore, we conduct the ablation studies on these two components. As reported in Tab. IV, PLCNet-base is the basic network without any component, yet achieves better precision than Deepsnake. Although equipping with the 1D lane attention module, the network attains no improvement on all metrics. This degradation may be caused by not considering patch-wise features to acquire local lane information. On the other hand, adding the patch feature extraction contributes to improved performance. After employing both two components, PLCNet obtains the best accuracy, which verifies that removing either of them will lead to inferior performance.

TABLE IV: Ablation studies on PLCNet. PLCNet-base means that the network does not use the 1D lane attention module or patch feature extraction.

| Methods | lane-IoU (%) ↑ | | | smooth-L1 ↓ (p) | L2 ↓ (p) | CD ↓ (p) |
|---|---|---|---|---|---|---|
| | 1 p | 2 ps | 3 ps | | | |
| Deepsnake | 21.9 | 51.1 | 65.0 | 0.148 | 0.574 | 1.339 |
| PLCNet-base | 22.2 | 52.5 | 66.6 | 0.043 | 0.276 | 1.224 |
| ✓ atten | 20.6 | 50.3 | 64.6 | 0.046 | 0.293 | 1.331 |
| ✓ patch | 22.8 | 53.8 | 67.8 | **0.037** | **0.256** | 1.149 |
| ✓ atten ✓ patch | **23.5** | **54.8** | **68.6** | 0.040 | 0.257 | **1.135** |

### G. Performance Evaluation

To evaluate the network utility, we present a comprehensive evaluation including the number of training epochs, inference speed, and GPU memory consumption in Tab. V. Owing to its one-stage architecture and the absence of evolution strategies, PLCNet needs fewer training iterations while delivering faster inference speed and lower GPU memory consumption than Deepsnake. Remarkably, PLCNet is capable of performing 1km lane correction within 0.3 second, significantly outpacing skilled annotators and proving its viability for practical implementation in autonomous driving projects.

TABLE V: The training and inference comparison results.

| Methods | Training epochs ↓ | Inference speed ↑ (fps) | GPU memory ↓ (MB) |
|---|---|---|---|
| Deepsnake | 150 | 6.9 | 1758 |
| PLCNet | **60** | **12.8** | **1747** |

## V. CONCLUSION

In this paper, we propose a patch-wise lane correction network (PLCNet) to automatically correct initial lane deviation for HD maps. PLCNet takes initial lane points and point could images as inputs and directly outputs point-wise correction offsets. To be specific, PLCNet first extract multi-scale lane features through a light-weight backbone. Then, it gathers patch-wise features centered at each initial lane point via ROIAlign to acquire local information. In addition, we design a 1D lane attention module to capture holistic lane features. Finally, a simple MLP is used to infer instance-level lane correction offsets. Our PLCNet supports local and global lane correction and the effectiveness is verified on a self-built dataset that covers urban, highway and factory scenes. In comparison with the contour-based Deepsnake, PLCNet achieves superior correction accuracy and expeditious inference speed. In future work, we aim to explore the automatic correction of polygon-shaped elements such as landmark and crosswalk, thereby demonstrating the extensibility of our proposed network.

## ACKNOWLEDGMENT

This work is supported by Uimobi Technology (Zhejiang) Co., LTD and Uisee Technology Co., LTD. We would like to thank all related colleagues in the SITU group for data pre-processing, HD map labeling and performance evaluation. Their efforts have significantly enhanced the quality of this work.

## REFERENCES

[1] Z. Bao, S. Hossain, H. Lang, and X. Lin, "High-Definition Map Generation Technologies for Autonomous Driving," *arXiv preprint arXiv:2206.05400*, 2022.
[2] D. Neven, B. De Brabandere, S. Georgoulis, M. Proesmans, and L. Van Gool, "Towards End-to-End Lane Detection: An Instance Segmentation Approach," in *IEEE Intelligent Vehicles Symposium, Proceedings*, pp. 286-291, 2018.
[3] L. Tabelini, R. Berriel, T. M. Paixão, C. Badue, A. F. de Souza, and T. Oliveira-Santos, "Keep your Eyes on the Lane: Real-time Attention-guided Lane Detection," *arXiv preprint arXiv:2010.12035*, 2020.
[4] Z. Feng, S. Guo, X. Tan, K. Xu, M. Wang, and L. Ma, "Rethinking Efficient Lane Detection via Curve Modeling," in *Proceedings of the IEEE Computer Society Conference on Computer Vision and Pattern Recognition*, pp. 17041–17049, 2022.
[5] Z. Xu, Y. Sun, and M. Liu, "Topo-boundary: A Benchmark Dataset on Topological Road-boundary Detection Using Aerial Images for Autonomous Driving," *IEEE Robotics and Automation Letters*, vol. 6, no .4, pp. 7248–7255, 2021.
[6] Z. Xu *et al.*, "CsBoundary: City-Scale Road-Boundary Detection in Aerial Images for High-Definition Maps," *IEEE Robotics and Automation Letters*, vol. 7, no .2, pp. 5063–5070, 2022.
[7] J. Liang, N. Homayounfar, W. C. Ma, S. Wang, and R. Urtasun, "Convolutional recurrent network for road boundary extraction," in *Proceedings of the IEEE Computer Society Conference on Computer Vision and Pattern Recognition*, pp. 9504–9513, 2019.
[8] Q. Li, Y. Wang, Y. Wang, and H. Zhao, "HDMapNet: An Online HD Map Construction and Evaluation Framework," *arXiv preprint arXiv:2107.06307*, 2021.
[9] Y. Liu, Y. Yuan, Y. Wang, Y. Wang, and H. Zhao, "VectorMapNet: End-to-end Vectorized HD Map Learning," *arXiv preprint arXiv:2206.08920*, 2022.
[10] D. E. Rumelhart, G. E. Hinton and R. J. Williams, "Learning representations by back-propagating errors," *Nature*, vol. 323, no. 6088, pp. 533-536, 1986.
[11] Y. LeCun, L. Bottou, Y. Bengio and P. Haffner, "Gradient-based learning applied to document recognition," *Proc. IEEE*, vol. 86, no. 11, pp. 2278-2324, Nov. 1998.
[12] V. Mnih, N. Heess, A. Graves, and K. Kavukcuoglu, "Recurrent models of visual attention," in *Advances in Neural Information Processing Systems*, vol. 3, pp. 2204-2212, 2014.
[13] S. Peng, W. Jiang, H. Pi, X. Li, H. Bao, and X. Zhou, "Deep Snake for Real-Time Instance Segmentation," *arXiv preprint arXiv:2001.01629*, 2020.
[14] T. F. Cootes, C. J. Taylor, D. H. Cooper, and J. Graham, "Active shape models - their training and application," *Computer Vision and Image Understanding*, vol. 61, no .1, pp. 38–59, 1995.
[15] A. H. Lang, S. Vora, H. Caesar, L. Zhou, J. Yang, and O. Beijbom, "Pointpillars: Fast encoders for object detection from point clouds," in *Proceedings of the IEEE Computer Society Conference on Computer Vision and Pattern Recognition*, pp. 12689–12697, 2019.
[16] W. Hess, D. Kohler, H. Rapp, and D. Andor, "Real-time loop closure in 2D LIDAR SLAM," in *Proceedings - IEEE International Conference on Robotics and Automation*, pp. 1271–1278, 2016.
[17] T. Shan and B. Englot, "LeGO-LOAM: Lightweight and Ground-Optimized Lidar Odometry and Mapping on Variable Terrain," in *IEEE International Conference on Intelligent Robots and Systems*, pp. 4758–4765, 2018.
[18] S. Sivaraman and M. M. Trivedi, "Integrated lane and vehicle detection, localization, and tracking: A synergistic approach," *IEEE Transactions on Intelligent Transportation Systems*, vol. 14, no .2, pp. 906–917, 2013.
[19] M. Tan and Q. V. Le, "EfficientNet: Rethinking model scaling for convolutional neural networks," *arXiv preprint arXiv:1905.11946*, 2019.
[20] K. He, G. Gkioxari, P. Dollar, and R. Girshick, "Mask R-CNN," in *Proceedings of the IEEE International Conference on Computer Vision*, pp. 2980–2988, 2017.
[21] D. P. Kingma and J. L. Ba, "Adam: A method for stochastic optimization," in *3rd International Conference on Learning Representations*, 2015.
[22] J. Deng, W. Dong, R. Socher, L.-J. Li, Kai Li, and Li Fei-Fei, "ImageNet: A large-scale hierarchical image database," in *Proceedings of the IEEE Computer Society Conference on Computer Vision and Pattern Recognition*, pp. 248-255, 2009.